\documentclass{elsarticle}

\usepackage{framed}
\usepackage{amssymb,amsmath}
\usepackage{graphicx}
\usepackage{color}
\usepackage[onehalfspacing]{setspace}

\usepackage{lineno,hyperref}
\modulolinenumbers[10000000]

\biboptions{authoryear}

\begin{document}

\begin{frontmatter}

\title{\vspace*{-0.8cm} On the use of Harrell's C for clinical risk prediction via random survival forests} 

\author[mymainaddress]{\vspace*{-0.2cm}Matthias Schmid$^{*,}$}
\cortext[mycorrespondingauthor]{Corresponding author, Email: matthias.schmid@imbie.uni-bonn.de,\\ \hspace*{.55cm}Phone: +49 228 287 15400\\
\hspace*{.25cm}$^{**}$Email: wright@imbs.uni-luebeck.de\\
\hspace*{.12cm}$^{***}$Email: ziegler@imbs.uni-luebeck.de}
\author[mysecondaryaddress]{Marvin N. Wright$^{**,}$}
\author[mysecondaryaddress,mythirdaddress,myfourthaddress,myfifthaddress]{Andreas Ziegler$^{***,}$}

\address[mymainaddress]{Institut f\"ur Medizinische Biometrie, Informatik und Epidemiologie, Rheinische 
	Friedrich-Wilhelms-Universit\"at Bonn, Sigmund-Freud-Str.~25, 53127 Bonn, Germany}
\address[mysecondaryaddress]{Institut f{\"u}r Medizinische Biometrie und Statistik, Universit{\"a}t zu L{\"u}beck, Universit{\"a}tsklinikum Schleswig-Holstein, Campus L{\"u}beck, Ratzeburger Allee 160, Geb. 24, 23562 L{\"u}beck, Germany}
\address[mythirdaddress]{Zentrum f{\"u}r Klinische Studien, Universit{\"a}t zu L{\"u}beck, Ratzeburger Allee 160, Geb. 24, 23562 L{\"u}beck, Germany}
\address[myfourthaddress]{School of Mathematics, Statistics and Computer Science, University of KwaZulu-Natal, King Edward Avenue, Scottsville 3209, Pietermaritzburg, South Africa}
\address[myfifthaddress]{Deutsches Zentrum f{\"u}r Herz-Kreislauf-Forschung, Standort Hamburg/Kiel/L{\"u}beck, Oudenarder Str.~16, 13347 Berlin, Germany\vspace{-0.8cm}}

\begin{abstract}
Random survival forests (RSF) are a powerful method for risk prediction of right-censored outcomes in biomedical research. RSF use the log-rank split criterion to form an ensemble of survival trees. The most common approach to evaluate the prediction accuracy of a RSF model is Harrell's concordance index for survival data (`$C$ index'). Conceptually, this strategy implies that the split criterion in RSF is different from the evaluation criterion of interest. This discrepancy can be overcome by using Harrell's $C$ for both node splitting and evaluation. We compare the difference between the two split criteria analytically and in simulation studies with respect to the preference of more unbalanced splits, termed end-cut preference (ECP). Specifically, we show that the log-rank statistic has a stronger ECP compared to the $C$ index. In simulation studies and with the help of two medical data sets we demonstrate that the accuracy of RSF predictions, as measured by Harrell's $C$, can be improved if the log-rank statistic is replaced by the $C$ index for node splitting. This is especially true in situations where the censoring rate or the fraction of informative continuous predictor variables is high. Conversely, log-rank splitting is preferable in noisy scenarios. Both $C$-based and log-rank splitting are implemented in the R~package \texttt{ranger}. We recommend Harrell's~$C$ as split criterion for use in smaller scale clinical studies and the log-rank split criterion for use in large-scale `omics' studies.
\end{abstract}

\begin{keyword}
concordance index \sep event history analysis \sep log-rank statistic \sep random survival forests \sep risk prediction \sep split rules
\end{keyword}

\end{frontmatter}

\linenumbers

\section*{Highlights}
\begin{itemize}
	\item Harrell's $C$ is proposed as a split criterion in random survival forests.
	\item Split points of continuous predictor variables differ substantially between Harrell's $C$ and log-rank splitting.
	\item The log-rank statistic has a stronger end-cut preference than Harrell's $C$.
	\item Harrell's $C$ outperforms log-rank splitting in smaller scale studies.
	\item Harrell's $C$ outperforms log-rank splitting if the censoring rate is high.
\end{itemize}

\section{Introduction}	\label{se:Introduction}

Random forests are among the most powerful methods for risk prediction in the biomedical sciences. The basic idea of random forests is to fit an ensemble of classification and regression trees (CART) to bootstrap samples that are generated from a set of learning data \citep{Breiman2001}. Ensemble predictions are obtained by averaging predictions from the individual trees \citep{Kruppa2014}. An important element of random forests is that only a small number of the predictor variables is made available for splitting, which is done at random in each node of a tree. With this randomization element, trees are decorrelated, and the variance of the ensemble prediction is reduced. The random selection of predictors also constitutes the main difference between random forests and earlier tree-based ensemble methods, such as bootstrap aggregating \citep[bagging; ][]{Breiman1996}.

Random forests were originally proposed for classifying dichotomous outcomes \citep{Breiman2001} and have been extended over the past $15$ years in a number of ways. For example, various methods have been developed for judging the importance of predictor variables, which may serve as a basis for variable selection \citep{Diaz-Uriarte2006,Ishwaran2011}. It is also possible to estimate individual probabilities for both dichotomous and categorical outcomes \citep{Kruppa2013} and to analyze continuous outcomes as well as right-censored event times \citep{Ishwaran2008}. Finally, considerable progress has been made in understanding the statistical properties of random forests, including results on consistency and asymptotic normality \citep{Biau2012, Arlot2014, Scornet2015, Wager2015, Wager2015a, Mentch2016}. Reviews can be found elsewhere; see, e.g., \cite{Boulesteix2012,Touw2013,Kruppa2014,Ziegler2014}.

The standard approach to analyze survival outcomes with random forests is termed {\em random survival forests} \cite[RSF; ][]{Ishwaran2008}. In RSF, an ensemble of survival trees is built, and tree splitting is performed by maximizing the log-rank statistic in each node. Ensemble predictions are given by averages over the cumulative hazard estimates in the terminal nodes of the trees, as estimated by the Nelson-Aalen estimator. The most common approach to evaluate the predictive performance of the ensemble is the calculation of the $C$ statistic for survival data, also termed `Harrell's $C$' \citep{Harrell1982, Ishwaran2008}. A value of $C = 0.5$ corresponds to a non-informative prediction rule whereas $C = 1$ corresponds to perfect association, implying that Harrell's $C$ is an easy-to-interpret coefficient that accounts for the whole range of the observed survival times \citep{Schmid2012}. In biomedical applications, in particular in the analysis of gene expression data, $C$ often ranges between the values $0.6$ and $0.75$. For example, estimates in this range were reported, among others, by \cite{VanBelleBioinf2011}, \cite{Schroeder2011} and \cite{Zhang2013}.
A remaining disadvantage of the RSF approach with $C$-based evaluation, however, is that the split criterion used for tree building is different from the performance criterion used to measure prediction accuracy. As a result, the performance measure of interest, i.e.\@ Harrell's $C$, may not be fully optimized by the log-rank splits and may even have characteristics that are not reflected by the log-rank statistic.

In this work, we therefore investigate whether the performance of RSF can be improved if Harrell's $C$ is used for {\em both} node splitting and the evaluation of prediction accuracy. In other words, the idea is to replace the log-rank split criterion by Harrell's $C$ and to determine split points that are optimal with respect to Harrell's $C$ in each node.
In Section \ref{se:Methods} we first provide a description of the random forest algorithm for survival data, which is followed by theoretical considerations on the two split criteria. In two simulation studies and with the re-analysis of two cancer data sets (Section \ref{se:Applications}) we finally demonstrate that the use of Harrell's $C$ can lead to systematic improvements in the predictive performance of RSF.

\newenvironment{einzug}{%
	\parskip2pt \parindent0pt 
	\def\lititem{\hangindent=0.5cm \hangafter1}}{%
	\par\ignorespaces} 
\newenvironment{einzug2}{%
	\parskip2pt \parindent0.5cm 
	\def\lititem{\hangindent=0.85cm \hangafter1}}{%
	\par\ignorespaces}
\newenvironment{einzug3}{%
	\parskip2pt \parindent0.5cm 
	\def\lititem{\hangindent=0.5cm \hangafter1}}{%
	\par\ignorespaces} 
	
\section{Methods}	\label{se:Methods} 

\subsection{Random survival forests}

\begin{figure}[t]
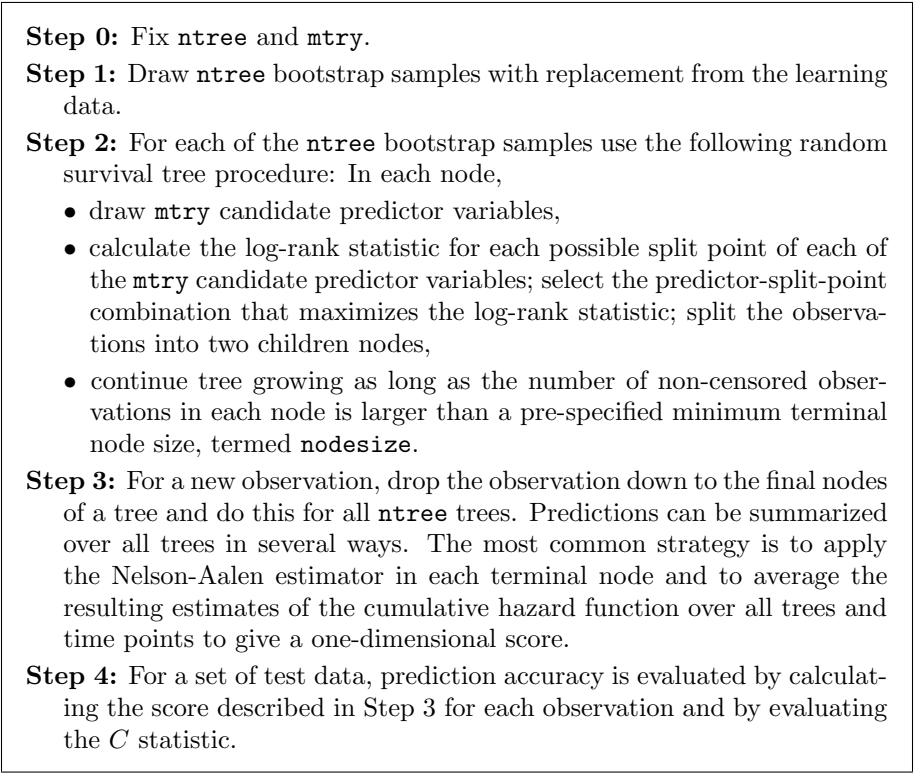

	\begin{framed}
		\begin{einzug}
			\lititem {\bf Step 0:} Fix \texttt{ntree} and \texttt{mtry}.
			
			\lititem{\bf  Step 1:} Draw \texttt{ntree} bootstrap samples with replacement from the learning data.
			
			\lititem {\bf  Step 2:} For each of the \texttt{ntree} bootstrap samples use the following random survival tree procedure: In each node,
		\end{einzug}
		\begin{einzug2}
			\lititem $\bullet$\hspace{0.05cm} draw \texttt{mtry} candidate predictor variables,
			
			\lititem $\bullet$\hspace{0.05cm} calculate the log-rank statistic for each possible split point of each of the \texttt{mtry} candidate predictor variables; select the predictor-split-point combination that maximizes the log-rank statistic; split the observations into two children nodes,
			
			\lititem $\bullet$\hspace{0.05cm} continue tree growing as long as the number of non-censored observations in each node is larger than a pre-specified minimum terminal node size, termed \texttt{nodesize}.
		\end{einzug2}
		
		\begin{einzug}
			\lititem {\bf Step 3:} For a new observation, drop the observation down to the final nodes of a tree and do this for all \texttt{ntree} trees. Predictions can be summarized over all trees in several ways. The most common strategy is to apply the Nelson-Aalen estimator in each terminal node and to average the resulting estimates of the cumulative hazard function over all trees and time points to give a one-dimensional score.
			
			\lititem {\bf  Step 4:} For a set of test data, prediction accuracy is evaluated by calculating the score described in Step 3 for each observation and by evaluating the $C$ statistic.
		\end{einzug}
	\end{framed}
	\caption{Schematic overview of the RSF algorithm proposed by \cite{Ishwaran2008}.} \label{Fig:Overview}
\end{figure}

Algorithm \ref{Fig:Overview} provides a description of the RSF algorithm for $n$ independent observations and $p$ predictor variables. Before the algorithm starts, the number of trees, termed \texttt{ntree}, of the RSF and the number of predictor variables \texttt{mtry} available for splitting at each node need to be defined. Recently, \cite{Lopes2015} derived the limiting distribution of the prediction error for dichotomous endpoints and showed how this finding may be used for determining optimized values of {\tt ntree}. \cite{Kruppa2013} demonstrated how the hyper-parameter \texttt{mtry} can be optimized.

An important feature of the RSF algorithm is the use of the log-rank statistic to split observations at each node and in every tree (Step $2$ in Figure \ref{Fig:Overview}). The log-rank statistic will be formally introduced below. At a specific node, the variable and the split point that maximize the log-rank statistic over all possible split points and all \texttt{mtry} variables are used for splitting. With this approach, the dissimilarity of the survival curves in the two children nodes is maximized. An alternative criterion for node splitting in Step $2$ is Harrell's $C$, which will also be considered below.

The performance of the random survival forest is evaluated using independent test data in Steps $3$ and $4$ of the algorithm. If no independent data are available, the out-of-bag data generated in Step 1 are used to evaluate the predictive performance. It is important to note that several summary measures are available in Step $3$ of the algorithm. For example, Kaplan-Meier or Nelson-Aalen estimates can be derived in each terminal node, and results may be averaged over all trees. In addition, confidence intervals can be obtained for these estimators \citep{Wager2014a,Wager2015,Mentch2016}.

\subsection{The C statistic and the log-rank statistic as split criteria} \label{s22}

In this section we introduce the use of Harrell's $C$ as split criterion, and we start with a theoretical analysis of both Harrell's $C$ and the log-rank statistic. Specifically, we show that both split criteria are special cases of the Gehan statistic \citep{Gehan1965} and that they can be obtained from the Gehan statistic by applying different standardization and weighting schemes. Since these schemes are different, differences can be expected between the two criteria regarding their splitting behavior in RSF. First, we introduce basic notation and provide formal definitions of the log-rank, $C$ and Gehan statistics. Second, we analyze the connection between the measures. Third, we provide a description of how Harrell's $C$ should be used as a split criterion in random forests for survival data.

\subsubsection*{Notation}
Throughout this paper we assume that RSF are fitted to a set of independent and identically distributed data of size $n$. The data are represented by vectors $( \tilde{T}_i, \Delta_i, X_{i1}, \ldots , X_{ip})$, $i=1,\ldots , n$, where $\tilde{T}_i$ is a possibly right-censored continuous survival time and $(X_{i1},\ldots , X_{ip})^\top$ is a vector of predictor variables. It is assumed that $\tilde{T}_i$ is the minimum of the true survival time~$T_i$ and an independent continuous censoring time $C_i$. The variable $\Delta_i := \mathrm{I}(T_i \le C_i)$ indicates whether $T_i$ has been fully observed ($\Delta_i = 1$) or not ($\Delta_i = 0$). To simplify notation, we assume that there are no tied observed survival times in the data. A predictor variable $X_j$, $j \in \{ 1, \ldots , p\}$, is called non-informative if the distribution of $\tilde{T}$ conditional on $X_j$ does not depend on $X_j$. Otherwise, $X_j$ is called informative.

The events are observed at $K$ ordered time points $t_{(1)} < \ldots < t_{(K)}$ with $K \le n$. The numbers of events and observations at risk at $t_{(k)}$, $k = 1,\ldots ,K$, are denoted by $d_k$ and $Y_k$, respectively.

As described in Step 3 of the RSF algorithm (Figure \ref{Fig:Overview}), the outcome of an RSF is calculated from the cumulative hazard estimates in the terminal nodes. A one-dimensional score $\eta_i \in\mathbb{R}$ is estimated for each observation $i = 1, \ldots, n$, by averaging the cumulative hazard estimates over all trees and all time points~$t_{(k)}$. 

\subsubsection*{Definition of Harrell's $C$} Harrell's $C$ \citep{Harrell1982} is given by
\begin{equation}	\label{harrellsc}
C = \frac{\sum_{i,j} \mathrm{I}(\tilde{T}_i > \tilde{T}_j) \cdot \mathrm{I}(\eta_j > \eta_i) \cdot \Delta_j }{\sum_{i,j} \mathrm{I}(\tilde{T}_i > \tilde{T}_j) \cdot \Delta_j} \, ,
\end{equation}
where the indices $i$ and $j$ refer to pairs of observations in the sample. The $C$~statistic is the number of concordant pairs of observations divided by the number of comparable pairs. Multiplication by the factor $\Delta_j$ in Eq.\ (\ref{harrellsc}) discards pairs of observations that are not comparable because the smaller survival time is censored, i.e., $\Delta_j = 0$.

Harrell's~$C$ is designed to estimate the concordance probability \linebreak $\mathrm{P}( \eta_j > \eta_i \,|\, T_i > T_j)$, which compares the rankings of two independent pairs of survival times $T_i, T_j$ and predictions $\eta_i, \eta_j$. The concordance probability evaluates whether large values of $\eta_i$ are associated with small values of $T_i$ and vice versa. Harrell' $C$ can also be interpreted as a summary measure of the area(s) under the time-dependent ROC curves \citep{Heagerty2005,Schmid2012}. A value of $C = 0.5$ corresponds to a non-informative prediction rule, whereas $C = 1$ corresponds to perfect association. It has thus been argued that Harrell's $C$ is an easy-to-interpret coefficient that accounts for the whole range of the observed survival times. Subsequently, \cite{Ishwaran2008} proposed to use Harrell's $C$ for evaluating the predictive performance of an RSF model.

In order to use Harrell's $C$ as a split criterion in RSF, it is necessary to define appropriate values of the score $\eta$. For this purpose, we assume that the observations in a node under consideration are split into two disjoint children nodes $\mathcal{G}_0$ and $\mathcal{G}_1$ according to the split point (or ``threshold'') of some candidate variable. Hence, to evaluate the goodness of the split using Harrell's $C$, we define $\gamma_i := \mathrm{I}( i \in \mathcal{G}_1) \in \{ 0,1 \}$ and estimate the concordance probability \linebreak $\mathrm{P}( \gamma_i < \gamma_j \,|\, T_i > T_j) = \mathrm{P}( i \in \mathcal{G}_0 , j \in \mathcal{G}_1 \,|\, T_i > T_j)$ by
\begin{eqnarray}
\label{c01}
C &=& \frac{\sum_{i,j} \mathrm{I}(\tilde{T}_i > \tilde{T}_j) \cdot \mathrm{I}(i \in \mathcal{G}_0 , j \in \mathcal{G}_1) \cdot \Delta_j }{\sum_{i \ne j} \mathrm{I}(\tilde{T}_i > \tilde{T}_j) \cdot \Delta_j} 
 \nonumber\\
&& + \
\frac{\sum_{i \ne j} 0.5 \cdot \mathrm{I}(\tilde{T}_i > \tilde{T}_j) \cdot \mathrm{I}(i \in \mathcal{G}_0 , j \in \mathcal{G}_0) \cdot \Delta_j }{\sum_{i \ne j} \mathrm{I}(\tilde{T}_i > \tilde{T}_j) \cdot \Delta_j} \nonumber\\
&&  +  \
\frac{\sum_{i \ne j} 0.5 \cdot \mathrm{I}(\tilde{T}_i > \tilde{T}_j) \cdot \mathrm{I}(i \in \mathcal{G}_1 , j \in \mathcal{G}_1) \cdot \Delta_j }{\sum_{i \ne j} \mathrm{I}(\tilde{T}_i > \tilde{T}_j) \cdot \Delta_j} \, .
\end{eqnarray}
The definition of (\ref{c01}) implies that a value of $0.5$ is assigned to pairs of observations belonging to the same child node.


\subsubsection*{The log-rank statistic}
The log-rank statistic is defined by
\begin{equation}
\label{logrank}
T_{\mathrm{log-rank}} = \frac{\left( \sum_k (d_{1k} - Y_{1k}\cdot d_{k}/Y_{k}) \right)^2 }
{\sum_k Y_{1k} Y_{0k} \cdot d_{k} (Y_{k} - d_{k}) / [Y_{k}^2 (Y_{k} - 1)]} \, ,
\end{equation}
where $d_{0k}, d_{1k}$ and $Y_{0k}, Y_{1k}$ refer to the numbers of events and observations at risk in groups $\mathcal{G}_0$ and ${G}_1$, respectively. It is a popular split criterion in survival trees \citep{LeBlanc1993} and has been adopted by \cite{Ishwaran2008} for use in RSF.

\subsubsection*{The Gehan statistic} The Gehan statistic \citep{Gehan1965} is given by
\begin{equation}
\label{gehan}
U = \sum_{i \in \mathcal{G}_0, j \in \mathcal{G}_1} \mathrm{I} (\tilde{T}_i > \tilde{T}_j) \cdot
\Delta_j \, - \, \sum_{i \in \mathcal{G}_0, j \in \mathcal{G}_1} \mathrm{I} (\tilde{T}_i < \tilde{T}_j) \cdot
\Delta_i \, .
\end{equation}
Assigning the value $1$ to pairs with $\tilde{T}_i > \tilde{T}_j$ and the value $-1$ to pairs with \linebreak $\tilde{T}_i < \tilde{T}_j$, the Gehan statistic evaluates whether survival times in $\mathcal{G}_0$ are systematically larger than those in $\mathcal{G}_1$. Pairs of observations are only considered if the smaller survival time is not censored.

\subsubsection*{Relationship between the Gehan statistic and the log-rank statistic}
Standardization and squaring of the Gehan statistic $U$ yields the Gehan-Wilcoxon statistic, which can be written as
\begin{equation}
\label{gehanWilcoxon}
T_{\mathrm{Gehan}} := \frac{U^2}{\mathrm{var}(U)} = \frac{\left( \sum_k Y_k \cdot (d_{1k} - Y_{1k}\cdot d_{k}/Y_{k}) \right)^2 }
{\sum_k Y_k^2 \cdot Y_{1k} Y_{0k} \cdot d_{k} (Y_{k} - d_{k}) / [Y_{k}^2 (Y_{k} - 1)]}
\end{equation}
\citep{Gehan1965,Tarone1977}.
Eq.\ (\ref{gehanWilcoxon}) shows that the Gehan-Wilcoxon statistic is a weighted version of the log-rank statistic $T_{\mathrm{log-rank}}$. Specifically, $T_{\mathrm{Gehan}}$ weights the summands in the numerator and denominator of $T_{\mathrm{log-rank}}$ by  the number of observations at risk, thereby assigning higher weight to events at early time points. This relationship gives rise to the more general class of Tarone-Ware test statistics \citep{Tarone1977}.

\subsubsection*{Relationship between the Gehan statistic and Harrell's $C$}
The Gehan statistic of Eq.\ (\ref{gehan}) can equivalently be written as
\begin{eqnarray}	\label{Gehanalt}
U &=& \sum_{i \in \mathcal{G}_0 , j \in \mathcal{G}_1} \mathrm{I} (\tilde{T}_i > \tilde{T}_j) \cdot
\Delta_j \, - \sum_{i \in \mathcal{G}_0 , j \in \mathcal{G}_1} \mathrm{I} (\tilde{T}_i < \tilde{T}_j) \cdot \Delta_i \nonumber\\ 
&=& 2 \, \cdot \sum_{i \in \mathcal{G}_0 , j \in \mathcal{G}_1} \mathrm{I} (\tilde{T}_i > \tilde{T}_j) \cdot
\Delta_j \, - \, N \, ,
\end{eqnarray}
where $N$ is the number of possible comparisons between $\mathcal{G}_0$ and $\mathcal{G}_1$. Similarly, the numerator of Harrell's $C$ in (\ref{c01}) can be written as
\begin{align}	\label{numalt}
\sum_{i \in \mathcal{G}_0 , j \in \mathcal{G}_1} \mathrm{I} (\tilde{T}_i > \tilde{T}_j) \cdot
\Delta_j \, + \, 0.5 \cdot N_0 + \, 0.5 \cdot N_1 \, ,
\end{align}
where $N_0$ and $N_1$ refer to the number of comparable pairs in groups $\mathcal{G}_0$ and~$\mathcal{G}_1$, respectively. Eqns.\ (\ref{Gehanalt}) and (\ref{numalt}) both depend on the term $ \sum_{i \in \mathcal{G}_0 , j \in \mathcal{G}_1} \mathrm{I} (\tilde{T}_i > \tilde{T}_j) \cdot \Delta_j$. As a consequence, Harrell's $C$ is linearly related to the Gehan statistic.

\subsubsection*{Conclusion about the relationships}
The above considerations show
\begin{enumerate}
	\item[(i)] that both Harrell's $C$ and log-rank statistic are based on the same goodness-of-split criterion, namely the Gehan statistic,
	\item[(ii)] that different standardization and weighting schemes are applied to transform the Gehan statistic into Harrell's $C$ and the log-rank statistic,
	\item[(iii)] that both measures include additional terms which become large if the groups $\mathcal{G}_0$ and $\mathcal{G}_1$ are  unbalanced. For $T_{\mathrm{log-rank}}$ this is seen by considering the denominator in Eq.\ (\ref{gehanWilcoxon}), which depends on the products $Y_{1k} \cdot Y_{0k}$. The latter term becomes small if $\mathcal{G}_0$ and $\mathcal{G}_1$ are unbalanced, effectively increasing $T_{\mathrm{Gehan}}$. It also increases $T_{\mathrm{log-rank}}$, whose denominator depends on the same products. In a similar way, Eq.\ (\ref{numalt}) shows that Harrell's~$C$ depends on the sum $N_0 + N_1$, which becomes large if $\mathcal{G}_0$ and $\mathcal{G}_1$ are unbalanced.
\end{enumerate}
The effect of these weighting schemes on RSF performance is investigated in Section \ref{se:Applications}.

\subsection{Use of Harrell's $C$ as split criterion in RSF}	\label{s23}

The definition of the binary values $\gamma_i = \mathrm{I}( i \in \mathcal{G}_1)$ allows for the following strategy to use Harrell's $C$ as split criterion in RSF: In Step 2 of the RSF algorithm (Figure \ref{Fig:Overview}), the $C$ statistic is evaluated in each node for all possible split points of all \texttt{mtry} candidate variables. In case $C < 0.5$, node labels are switched and Harrell's $C$ is replaced by $1 - C$. Based on the obtained set of $C$ values, the predictor-split-point combination maximizing Harrell's $C$ is used to split the observations into the children nodes $\mathcal{G}_0$ and $\mathcal{G}_1$. To account for the fact that Harrell's $C$ discards pairs of observations with censored smaller observed time, which is known to result in an upward-bias in the estimation of the concordance probability $\mathrm{P}( i \in \mathcal{G}_0 , j \in \mathcal{G}_1 \,|\, T_i > T_j)$ \citep{Schmid2012}, one could additionally use the estimator proposed by \cite{Uno2011} (`Uno's $C$') for node splitting. Uno's $C$ ensures consistent estimation of the concordance probability by inverse probability-of-censoring (IPC) weighting of the terms in (\ref{c01}). Here, we do not propose to replace Harrell's $C$ by Uno's $C$, as this strategy requires the additional estimation of the IPC weights in each node. Apart from the computational cost, this estimation step is not feasible in deeply grown trees with small minimum node size. Furthermore, we demonstrate in simulation study 2 that the conclusions are not altered when Harrell's $C$ is used instead of Uno's $C$.

RSF with log-rank-based and (Harrell's-)$C$-based splitting are implemented in the R package \texttt{ranger} (\citealp{Wright2016}, options \texttt{splitrule = "C"} and \texttt{splitrule = "logrank"}). Default values of $\texttt{ntree}$ and $\texttt{mtry}$ are $500$ and~$\sqrt{p}$, respectively, where $p$ is the number of predictor variables.

\subsection{Simulation study 1 -- threshold selection}	\label{s31}

We compared the performance of the two split criteria by carrying out two simulation studies. In simulation study 1, we investigated the behavior of the log-rank and $C$ statistics with regard to threshold selection. The purpose of this study was to show that the thresholds of a given predictor variable that maximize the two statistics systematically differ. Specifically, the log-rank statistic has a stronger tendency to generate unbalanced children nodes than Harrell's $C$. In simulation study 2, we investigated the use of Harrell's $C$ and the log-rank statistic as split criteria in RSF and analyzed the prediction accuracy of the resulting model fits.

To study differences in the threshold selection between the two split criteria, we first considered a single continuous predictor variable $x$ that followed a uniform distribution on $[-3,3]$ (simulation study 1.a)). The sample size in simulation study 1.a) was set to $n = 1000$ independent observations. Survival times were generated from a standard exponential distribution without taking the values of $x$ into account. Consequently, simulation study 1.a) reflected the situation where a non-informative predictor is considered for node splitting. Note that the use of a uniformly distributed predictor variable $x$ corresponds to a very general scenario because every continuous predictor variable can be transformed into a uniformly distributed continuous variable by use of its distribution function. In particular, this transformation does not affect the splitting behavior of the two split criteria Harrell's $C$ and log-rank statistic. Censoring times were generated from an exponential distribution whose rate was adjusted such that $50\%$ of the observed survival times were right-censored.

In simulation study 1.b) we set the sample size to $n=100$ and investigated the tendency of the two split criteria to generate unbalanced children nodes in settings with an informative predictor variable. Specifically, we investigated the threshold models $T = \exp ( \mathrm{I}(x > 0.25) + \varepsilon )$ and $T = \exp ( \mathrm{I}(x > 0.75) + \varepsilon )$, with $x$ and $\varepsilon$ following independent standard uniform and normal distributions, respectively. For both threshold models, three censoring rates ($25\%$, $50\%$ and~$75\%$) were considered. Censoring times were generated in the same way as before. In both parts of simulation study~1, $1000$ replications were used.

\subsection{Simulation study 2 -- predictive performance of RSF} \label{s32}

To study the effects of $C$-based and log-rank-based splitting on the performance of RSF, we based the data-generating model on four informative predictor variables $x_1, \ldots, x_4$ that followed a multivariate standard normal distribution with pairwise correlation $\rho = 0.5$. After random number generation, the values of $x_1, \ldots, x_4$ were rounded to multiples of $0.1$. To investigate the effect of many noise variables, we also considered models with $p=10$, $p=505$, and $p=1005$ predictor variables. This was done by adding non-informative variables to the data sets that followed the same distribution as $x_1, \ldots, x_4$. Two sample sizes were considered in this simulation study ($n = 100, 300$).

\begin{figure}[t!]
	\centering 
	\includegraphics[scale = 0.6]{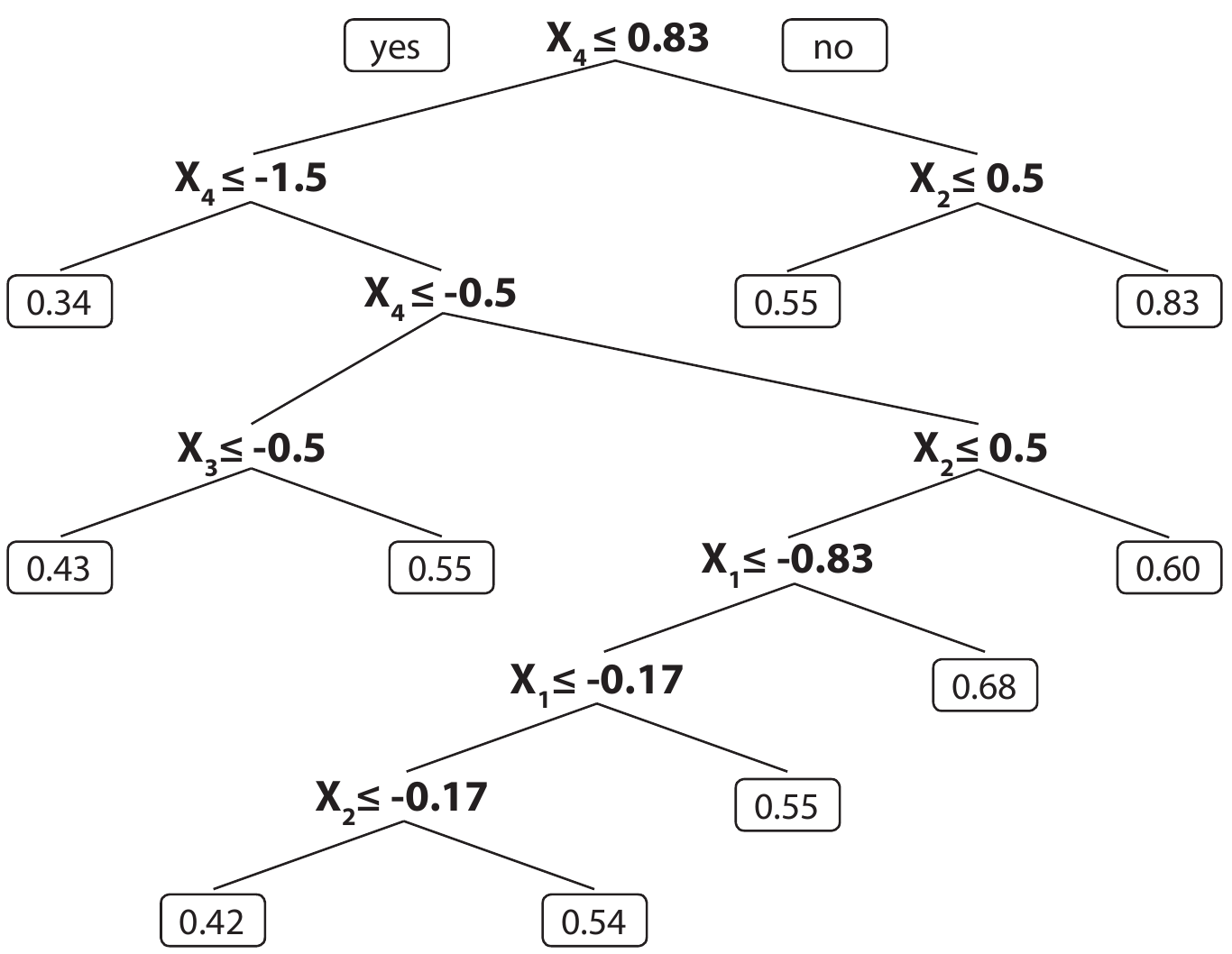} 
	\caption{Probability tree model to generate the location parameters $\lambda_i$ in simulation study~2. The numbers in the terminal nodes refer to the values of $\lambda_i$ that were obtained by dropping the values of the predictor variables $x_1, \ldots, x_4$ down the tree.} \label{Fig:Tree}
\end{figure}

To generate the observed survival times, we first dropped the values of the predictor variables down a probability tree model with fixed split rules and thresholds (Figure \ref{Fig:Tree}). This procedure was equivalent to the application of a non-random function of the predictor variables and resulted in a set of one-dimensional location parameters $\lambda_i \in [0,1]$, $i=1, \ldots, n$. Next, the $\lambda_i$ values were incorporated into the Weibull model $\ln (T_i) = \lambda_i + \sigma \cdot \varepsilon_i$ from which the survival times $T_i$ were generated. The noise variable $\varepsilon_i$ was chosen independently of $\lambda_i$ and followed a standard extreme value distribution. The parameter $\sigma$ quantified the amount of noise added to the location parameter $\lambda_i$. It was adjusted such that the empirical standard deviation of $\gamma_i$ was equal to~$\sigma$. This combination of $\lambda_i$ and the Weibull model resulted in a data-generating process that was a combination of a tree model and a proportional hazards model. We considered three censoring rates ($25\%$, $50\%$, $75\%$). Censoring times were generated as before (simulation study 1).

In simulation study 2.a), the true values of the predictor variables were used for RSF estimation. In simulation study 2.b), we dichotomized the predictor variables $\mathrm{I}(x_j > 0)$, $j=1, \ldots, p$, and investigated the behavior of log-rank and $C$-based splitting in the presence of categorical predictors. The tree model defined in Figure \ref{Fig:Tree} was again used generate the set of values $\lambda_i \in [0,1]$, \linebreak $i=1, \ldots, n$.

RSF were estimated using the \texttt{ranger} package with \texttt{ntree} $ = 500$ and minimum terminal \texttt{nodesize} $ = 3$. The other RSF parameters were chosen by default values. Prediction accuracy was measured by Harrell's $C$. 
In simulation study~2.a) we also investigated the performance of RSF when Uno's $C$ was used for evaluation. Both measures were evaluated on $500$ independent test data sets of size $n=1000$ each. The censoring rates in the test data sets were the same as those in the corresponding learning data sets.

\subsection{Data analysis 1 -- gene expression data to predict the time to distant metastases in breast cancer}
In the first data analysis, we applied RSF to gene expression data for metas-tasis-free survival in patients with node-negative breast cancer. Data were originally collected by \cite{Desmedt2007} to evaluate a 76-gene expression signature derived from Affymetrix microarrays \citep{Wang2005} in $198$ patients. The outcome variable was the time from diagnosis to distant metastases. In addition to the expression levels of the $76$ genes we considered the five clinical variables estrogen receptor status (positive/negative), tumor size (mm), tumor grade (poor/indermediate/good differentiation), age (years) and center (five hospitals, see \citealp{Desmedt2007} for details). The data are publicly available as part of the Gene Expression Omnibus (GEO) database (accession number GSE7390). Two observations with missing information on tumor grade were dropped from statistical analysis. Following the strategy by \cite{Desmedt2007}, observed survival times were censored at $10$ years. Observed metastasis-free survival times then ranged from $125$ days to $3652$ days, with $79.08$\% of the survival times being censored.

In the first step, we considered the five clinical variables only and fitted RSF to subsamples of sizes $\frac{1}{3}$ ($n = 64$), $\frac{1}{2}$ ($n = 97$) and $\frac{2}{3}$ ($n = 130$) of the $196$ observations. For each of the three sample sizes, $100$ random subsamples were generated. RSF predictions were evaluated on $100$ subsamples of size $66$ that were not part of the learning data sets. The same evaluation data could thus be used for all three sizes of the learning data. Both Harrell's $C$ and the log-rank statistic were used for node splitting. The number of trees was set to \texttt{ntree}$= 500$. For all tuning parameters, we used default values.

In the second step, we fitted RSF to the same $100$ subsamples, this time considering both the clinical variables and the expression values of the $76$ genes. Because all $76$ gene expression variables were identified as being predictive for the time to distant metastases \citep{Wang2005}, it could be assumed that all learning data sets were free of noise variables.

In the third step, we repeated the resampling analysis, this time adding various numbers of non-informative variables ($p_{\mathrm{add}} = 1000, 1500, 2000, \ldots , 5000$) to the clinical variables and the gene expression variables. All non-informative variables were generated from a multivariate standard normal distribution with pairwise correlation $\rho = 0.5$. This last setting allows to compare the RSF results with the findings from the high-dimensional simulation scenario in simulation study 2.b).


\subsection{Data analysis 2 -- survival in patients with diffuse large B-cell lymphoma}
In the second data analysis, we analyzed survival in patients with diffuse large B-cell lymphoma \citep{Rosenwald2002} using RSF. The aim of our analysis was to investigate the performance of the two split criteria in a higher-dimensional scenario where the signal-to-noise ratio of the predictors was unknown. Data were originally collected by \cite{Rosenwald2002} and comprised $p = 7399$ gene expression values from Lymphochip DNA microarrays. Here, we consider a subsample of $n=240$ patients that was analyzed previously by \cite{Binder2008}. The outcome was overall survival after chemotherapy. The median follow-up time was $2.8$ years; $42.5$\% of the survival times were censored. Because some of the microarray features contained missing values, we used the imputation method of \cite{Schumacher2007}.

We applied $C$-based and log-rank-based splitting as before, and RSF analysis was based on the same resampling approach as above. Specifically, we fitted RSF models to 100 subsamples of the original data using sample sizes of $\frac{1}{3}$ ($n = 80$), $\frac{1}{2}$ ($n = 120$) and $\frac{2}{3}$ ($n = 160$) of the $240$ observations.
RSF predictions were evaluated on $100$ subsamples of size $80$ that were not part of the respective learning data sets.


\section{Results}	\label{se:Applications}

\subsection{Simulation study 1 -- threshold selection}
\begin{figure}[t!]
	\centering \vspace{-0.6cm}
	\includegraphics[scale = 0.6]{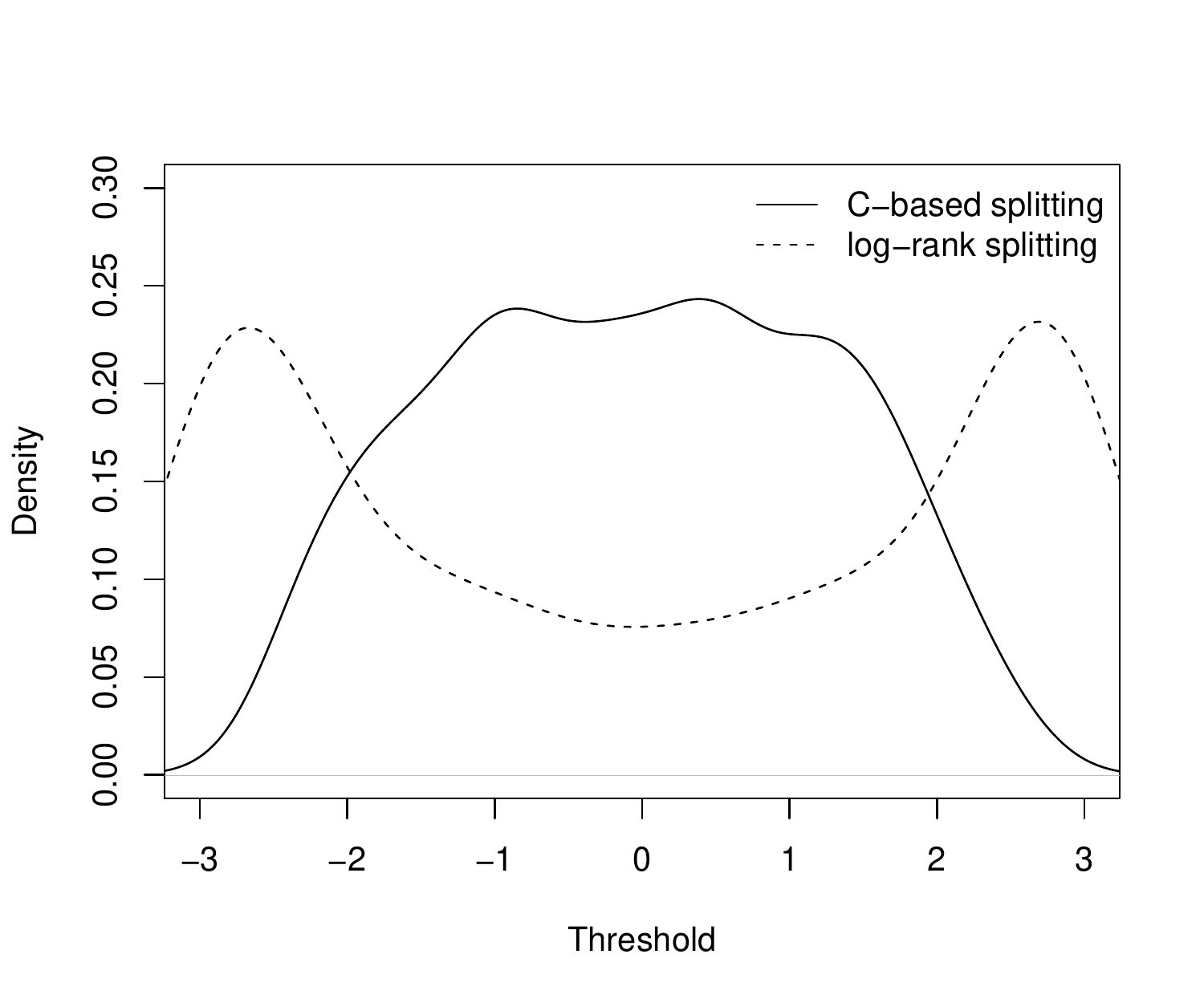} \vspace{-.3cm}
	\caption{Kernel density plots of the thresholds that were selected by the $C$ and log-rank statistics in simulation study 1.a). The thresholds refer to a non-informative predictor variable~$x$ that was uniformly distributed on $[-3,3]$. The kernel density plots therefore imply that Harrell's $C$ resulted in more balanced groupings of the observations than the log-rank statistic.}  \label{Fig:Threshold1}
\end{figure}

Figure \ref{Fig:Threshold1} shows that the selected thresholds differed substantially for the two split criteria when a uniformly distributed predictor variable with no explanatory value was used. Specifically, the kernel density plot of the thresholds selected by Harrell's $C$ reaches its maximum near the midpoint of the interval~$[-3,3]$, whereas the corresponding density plot for the thresholds selected by the log-rank statistic is characterized by two distinct peaks near the boundary values $-3$ and $3$. This implies that Harrell's $C$ resulted in more balanced groupings of the observations than the log-rank statistic.

\begin{figure}[t!]
	\centering \vspace{-0.5cm}
	\includegraphics[scale = 0.42]{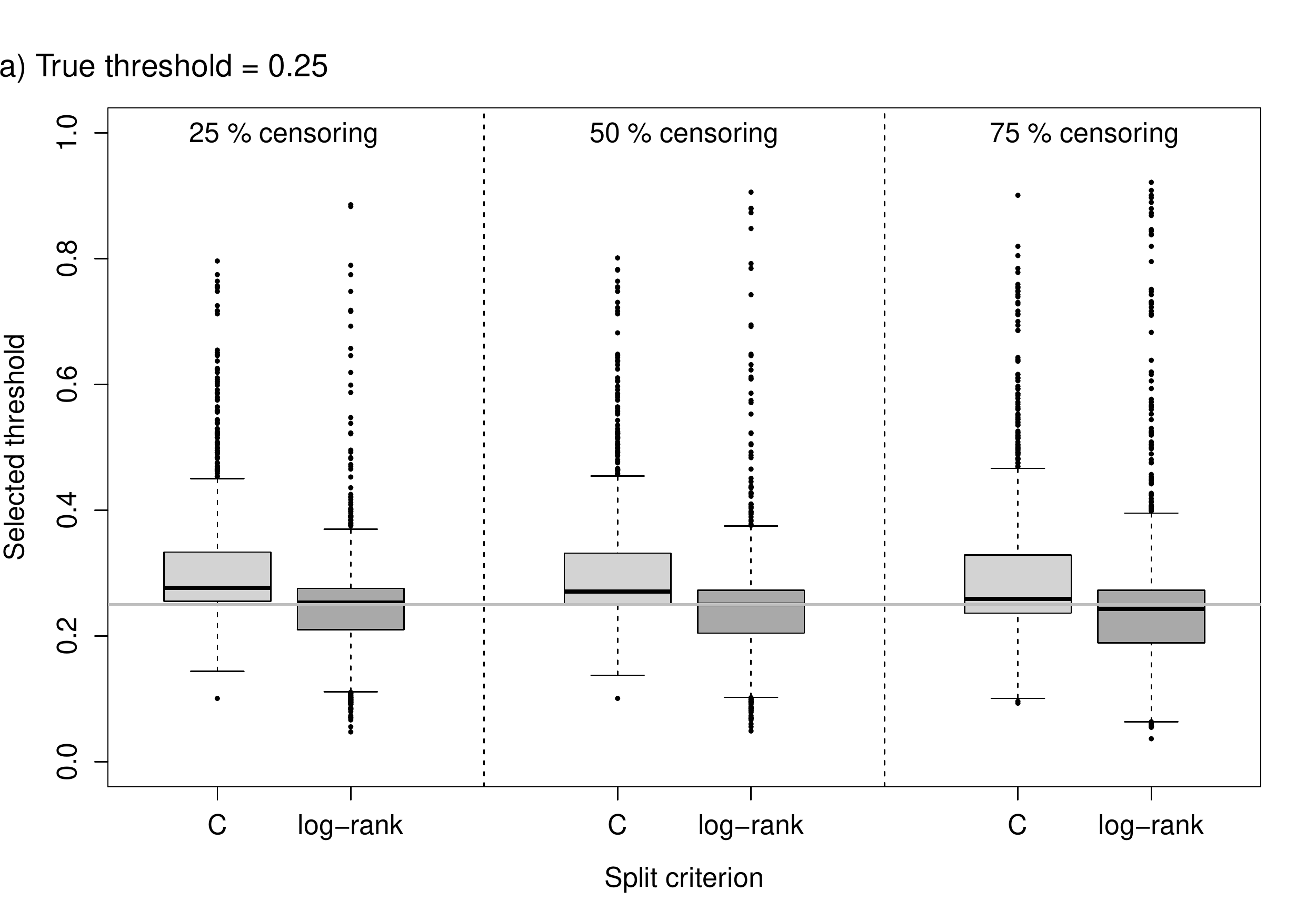}\vspace{-.3cm}
	\includegraphics[scale = 0.42]{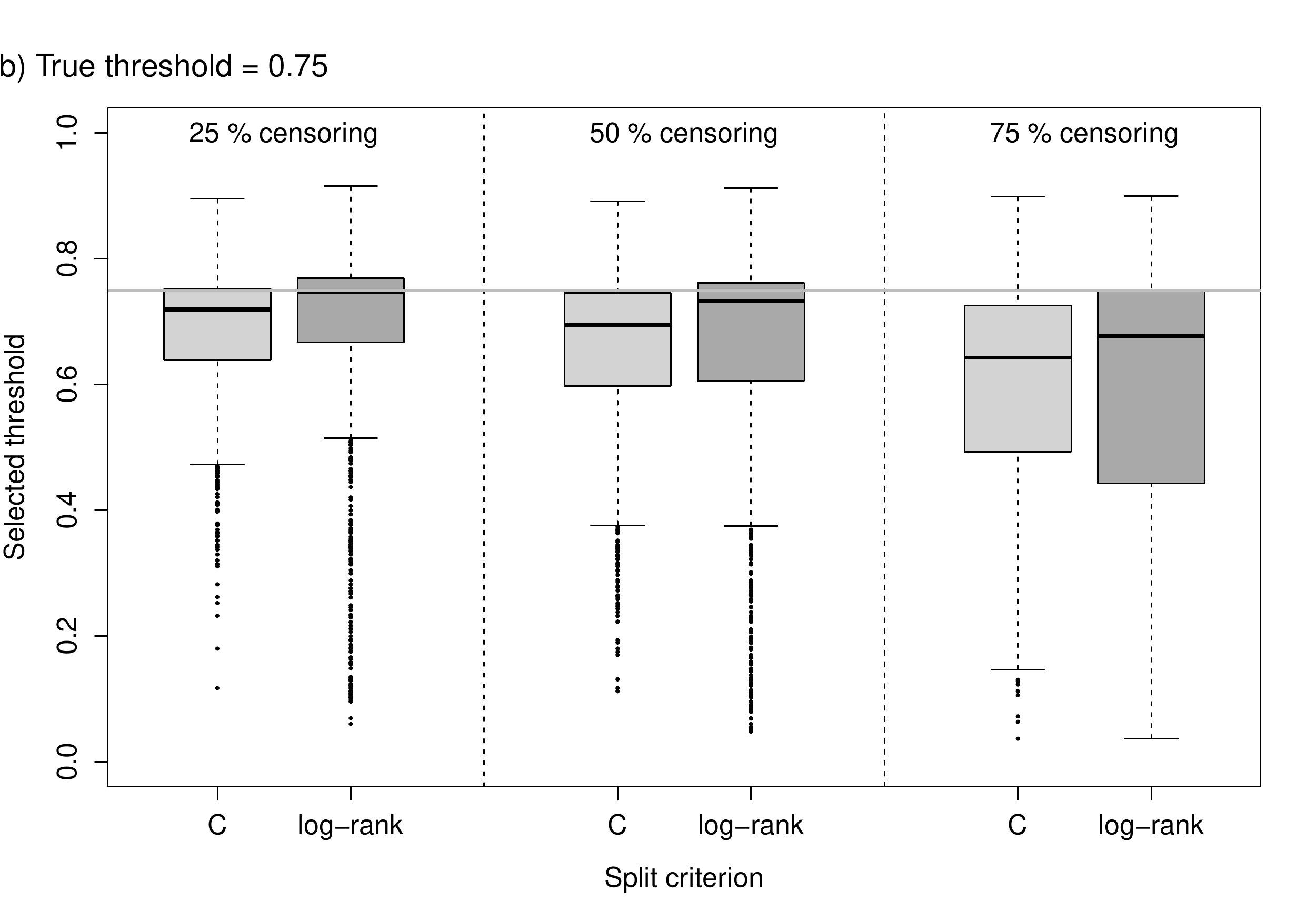}\vspace{-.2cm}
	\caption{Boxplots of the thresholds optimizing the $C$ and log-rank statistics in simulation study 1.b). The true thresholds were $0.25$ and $0.75$. On average, threshold values from the $C$ statistic were closer to $0.5$ when compared with the log-rank statistic. Thus, $C$-based splitting resulted in more balanced children nodes than log-rank splitting.} \label{Fig:Threshold2a}
\end{figure}

Simulation study 1.b) confirms the observation that Harrell's $C$ resulted in more balanced children nodes than the log-rank statistic, as both the median threshold values and the whole empirical threshold distributions obtained from the $C$ statistic were throughout closer to $0.5$ than those obtained from the log-rank statistic (Figure \ref{Fig:Threshold2a}).

\subsection{Simulation study 2 -- predictive performance of RSF}
\label{subs:sim2}
\begin{figure}[t!]
	\centering \vspace{-0.7cm}
	\includegraphics[scale = 0.33]{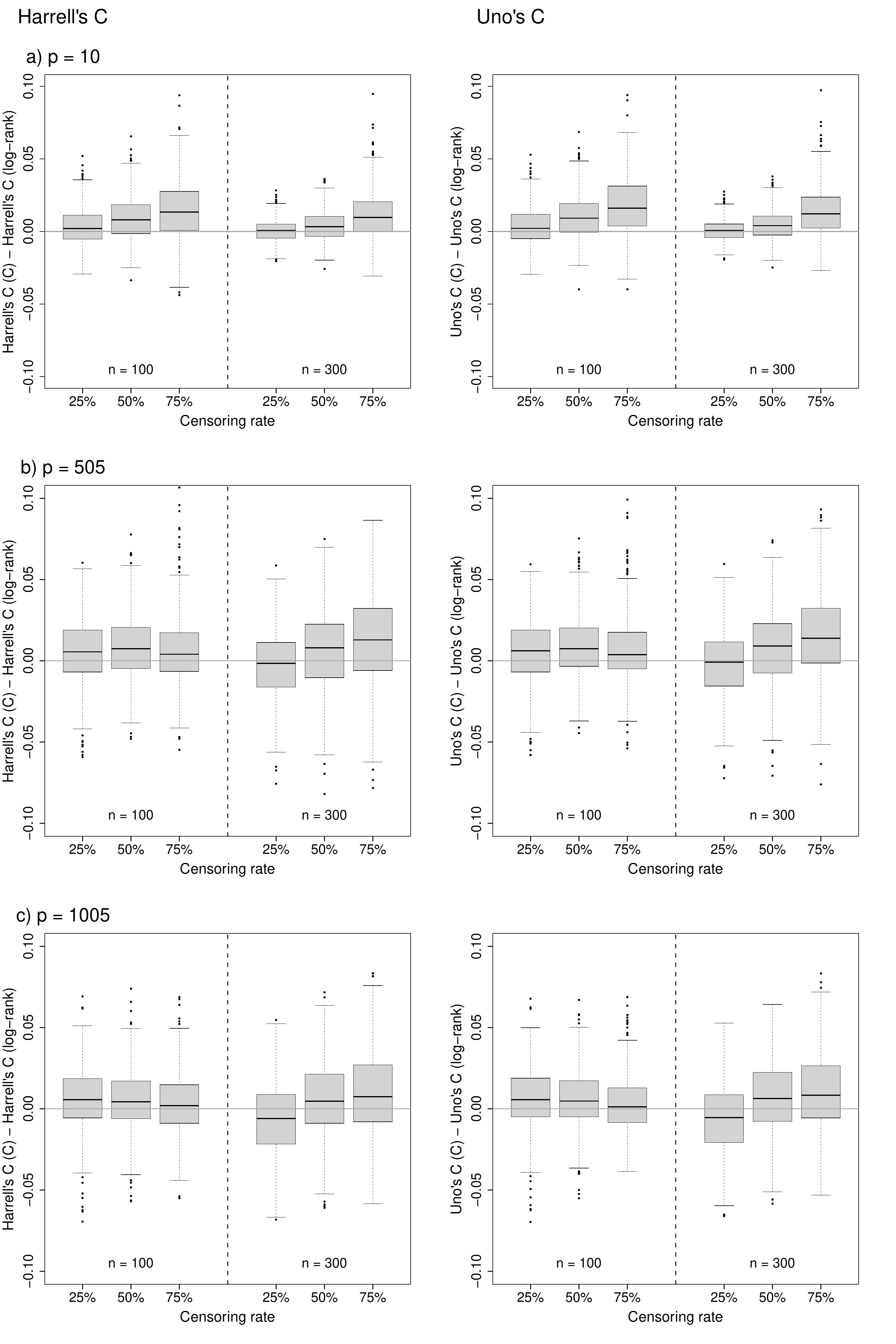}
	\caption{Results of simulation study 2.a). The panels show boxplots of the performance differences between $C$-based and log-rank-based splitting in RSF, as measured by Harrell's~$C$ (left column) and Uno's $C$ (right column) using independent test data. $C$-based splitting performed better than log-rank-based splitting if the difference was $> 0$.}
		\label{fig:simuEval}
\end{figure}

The panels in the left column of Figure \ref{fig:simuEval} (simulation study 2.a)) display the differences in RSF prediction accuracy between $C$-based and log-rank-based splitting, as measured by Harrell's $C$. The panels in the right column show the differences in RSF prediction accuracy when Uno's $C$ was used to evaluate predictive performance instead of Harrell's $C$. Results obtained from Harrel's~$C$ and Uno's $C$ were very similar throughout.

All panels in Figure \ref{fig:simuEval} show systematic differences between $C$-based and log-rank-based splitting. In most scenarios, $C$-based splitting improved the performance of RSF, which is indicated by an upward shift of the boxplots  in Figure \ref{fig:simuEval} and by median values lying above the horizontal line at zero. The magnitude of the observed differences depended on the censoring rate, with higher censoring rates tending to result in an improved performance of $C$-based splitting relative to log-rank-based splitting (see in particular Figure \ref{fig:simuEval}.a)). Increasing the number of non-informative variables tended to improve the performance of the log-rank split criterion relative to Harrell's~$C$. For example, the performance of log-rank based RSF was better for $1001$ non-informative predictors (Figure~\ref{fig:simuEval}.c)) than it was for six non-informative predictors (Figure \ref{fig:simuEval}.a)). In the latter figure, most of the boxplots are characterized by a distinct upward shift, whereas most of the medians presented in the former figure are only slightly larger than zero.

\begin{figure}[t!]
	\centering \vspace{-.7cm} \hspace{-0.45cm}
	\includegraphics[scale = 0.32]{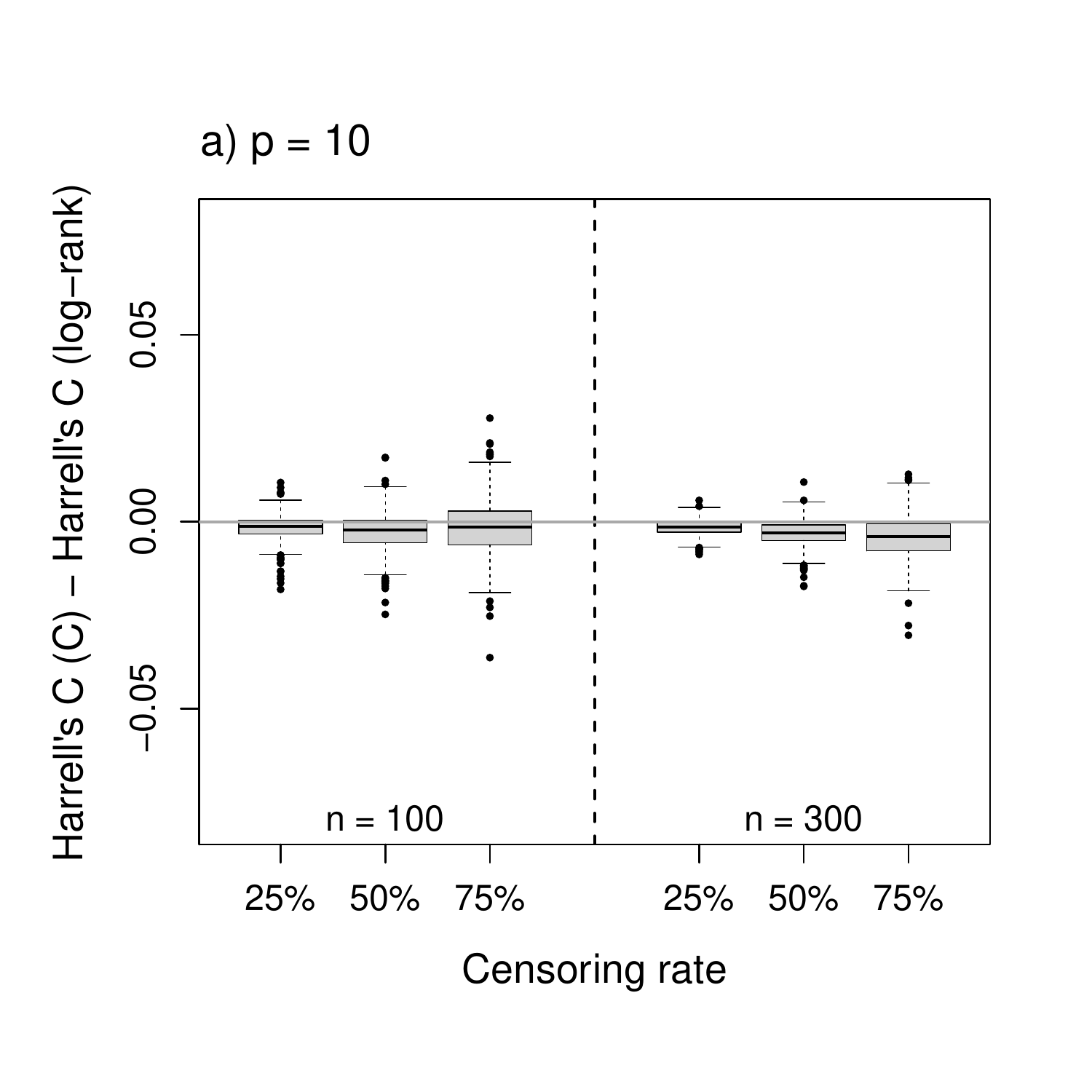}\hspace{-1.2cm}
	\includegraphics[scale = 0.32]{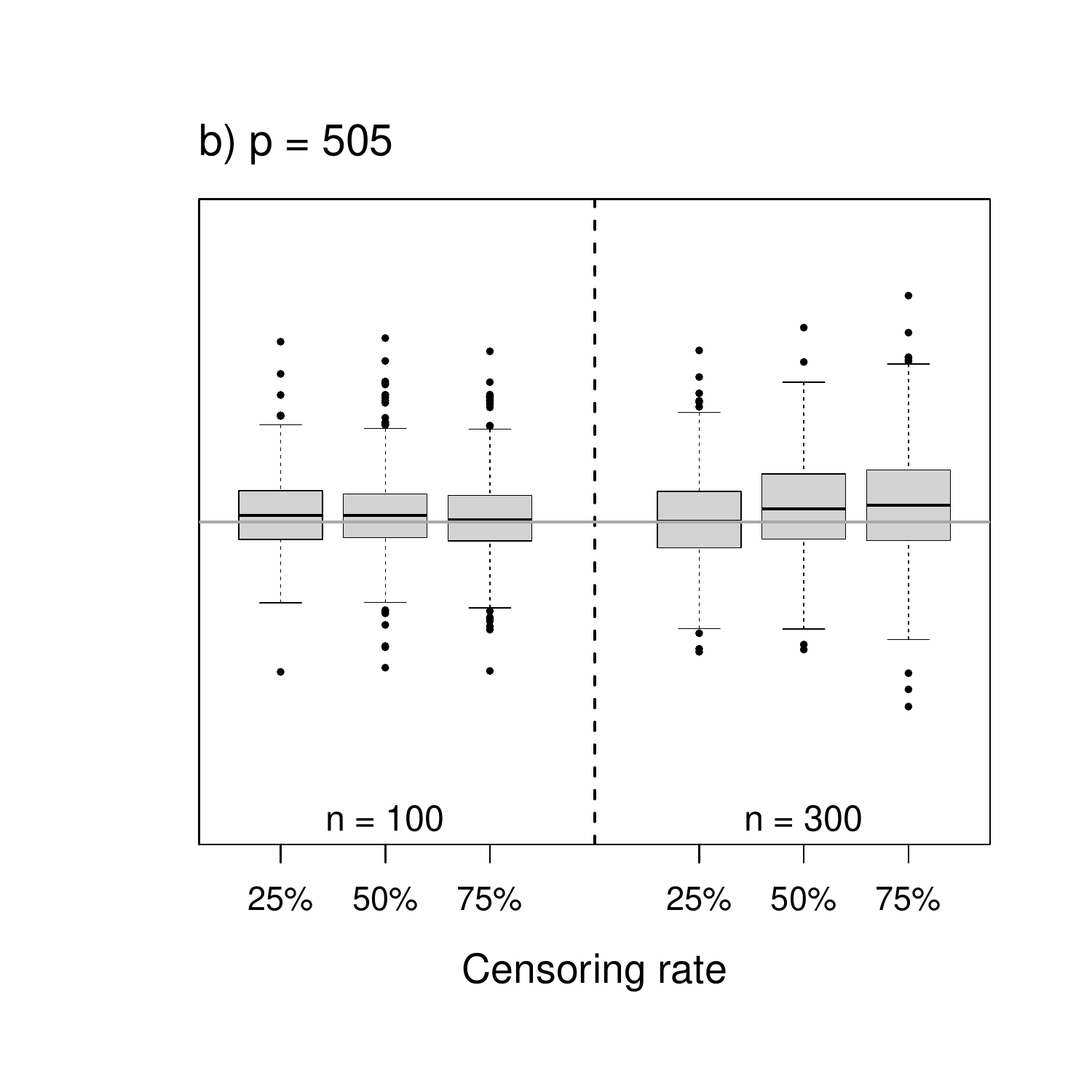}\hspace{-1.2cm}
	\includegraphics[scale = 0.32]{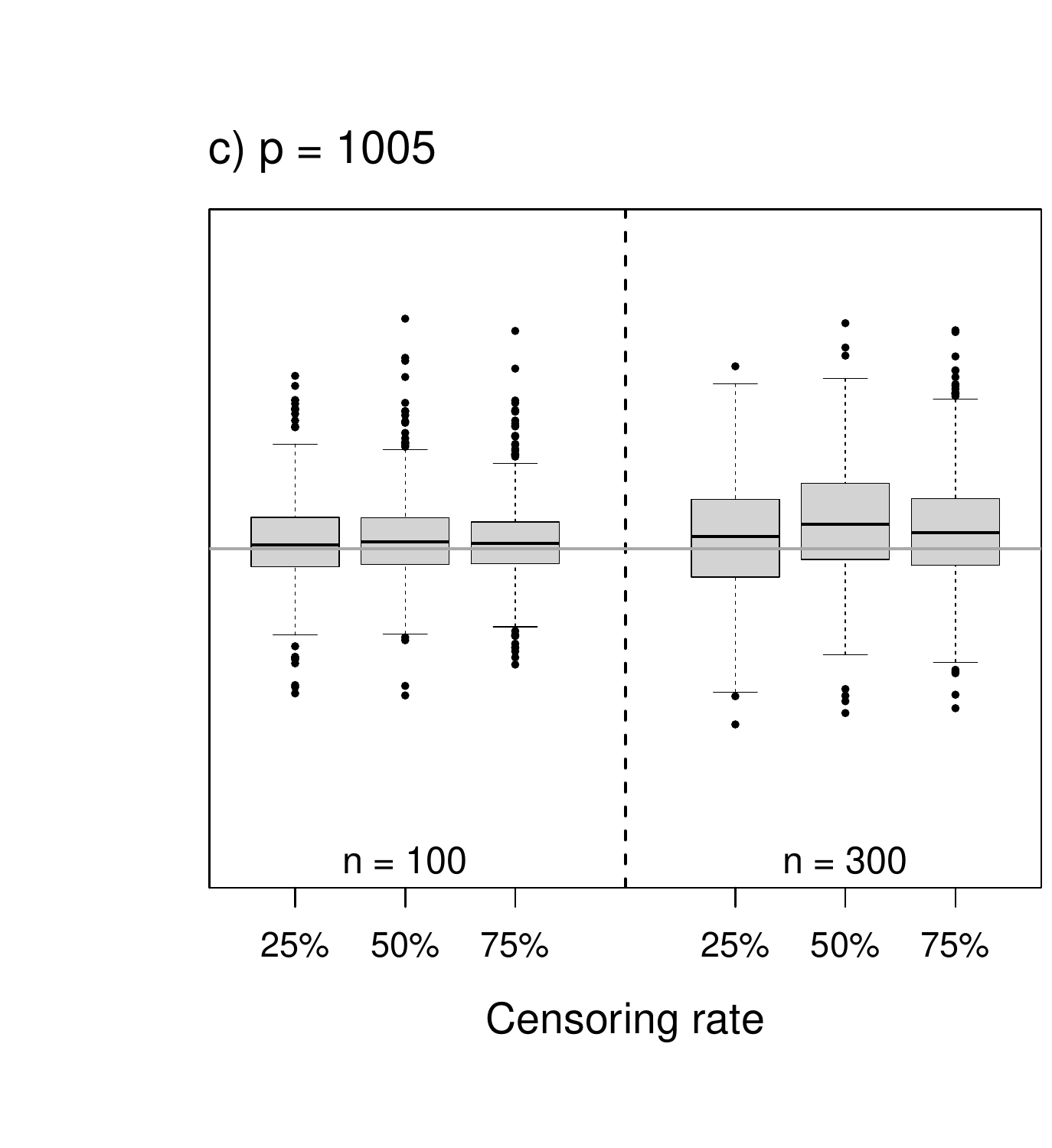} 
	\caption{Results of simulation study 2.b). The panels show boxplots of the performance differences between $C$-based and log-rank-based splitting in RSF, as measured by Harrell's~$C$ using independent test data. Predictor variables were dichotomized, as described in Section~\ref{subs:sim2}. $C$-based splitting performed better than log-rank-based splitting if the difference was $> 0$.} \label{fig:simuEvalDiscrete}
\end{figure}

Figure \ref{fig:simuEvalDiscrete} shows the differences in RSF prediction accuracy that were obtained from the dichotomized predictor variables $\mathrm{I}(X_j > 0)$, $j=1, \ldots , p$ (simulation study 2.b)). Apart from the low-dimensional scenario with $p = 10$, where log-rank splitting outperformed $C$-based splitting, no systematic differences were observed between $C$-based and log-rank-based splitting. In fact, most of the boxplots in Figure \ref{fig:simuEvalDiscrete} are either close to zero or are centered around the horizontal line at zero. While substantial differences were found for the split points of $C$-based and log-rank splitting in the simulation studies with continuous predictor variables (see in particular Figure~\ref{Fig:Threshold1}, and also the upward shifts in Figure~\ref{fig:simuEval}.a)), split rules became very similar when predictor variables were dichotomized (Figure~\ref{fig:simuEvalDiscrete}).

\subsection{Data analysis 1 -- gene expression data to predict the time to distant metastases in breast cancer}	\label{s41}

\begin{figure}[t!]
	\centering \vspace{-.5cm}
	\includegraphics[scale = 0.42]{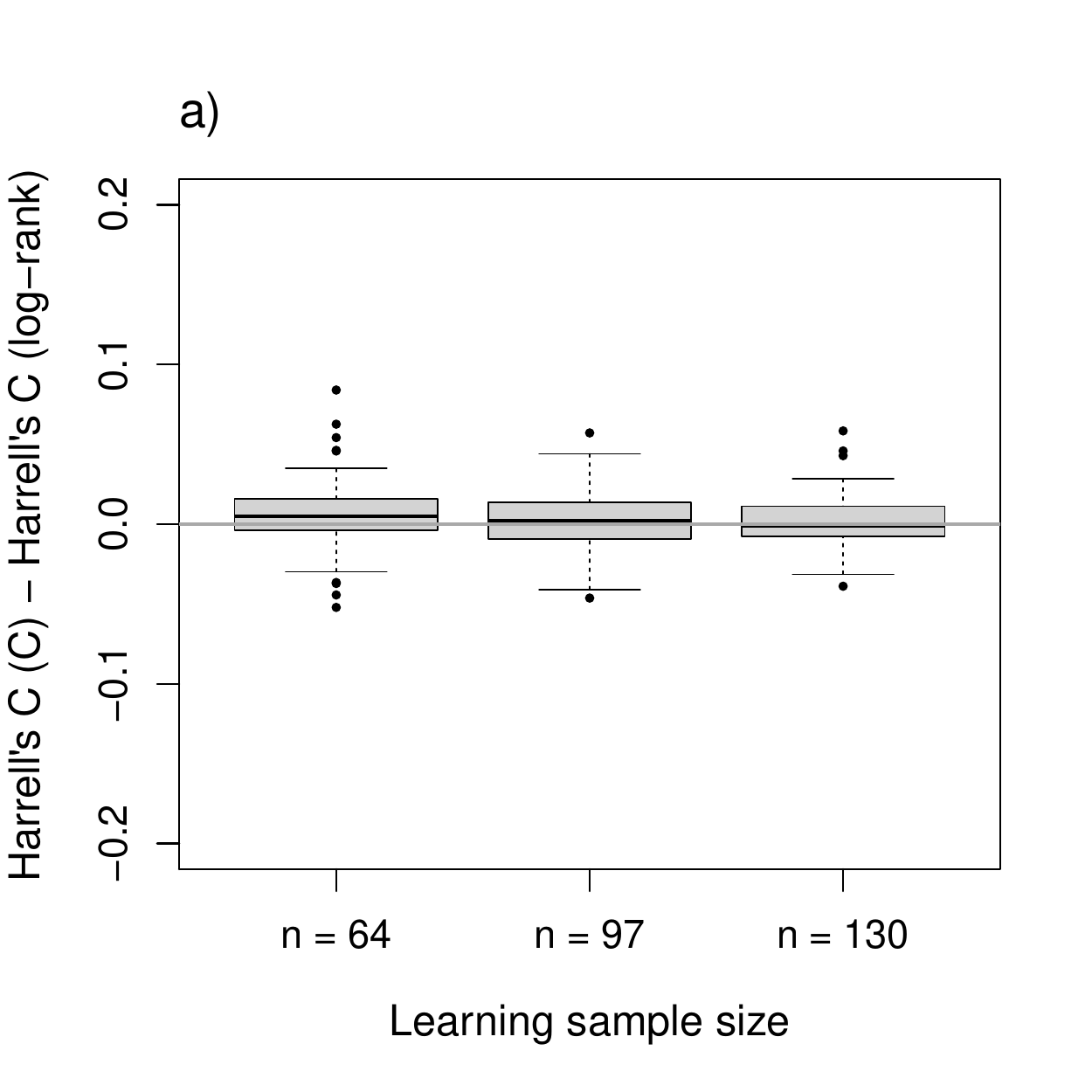}\hspace{-0.5cm}
	\includegraphics[scale = 0.42]{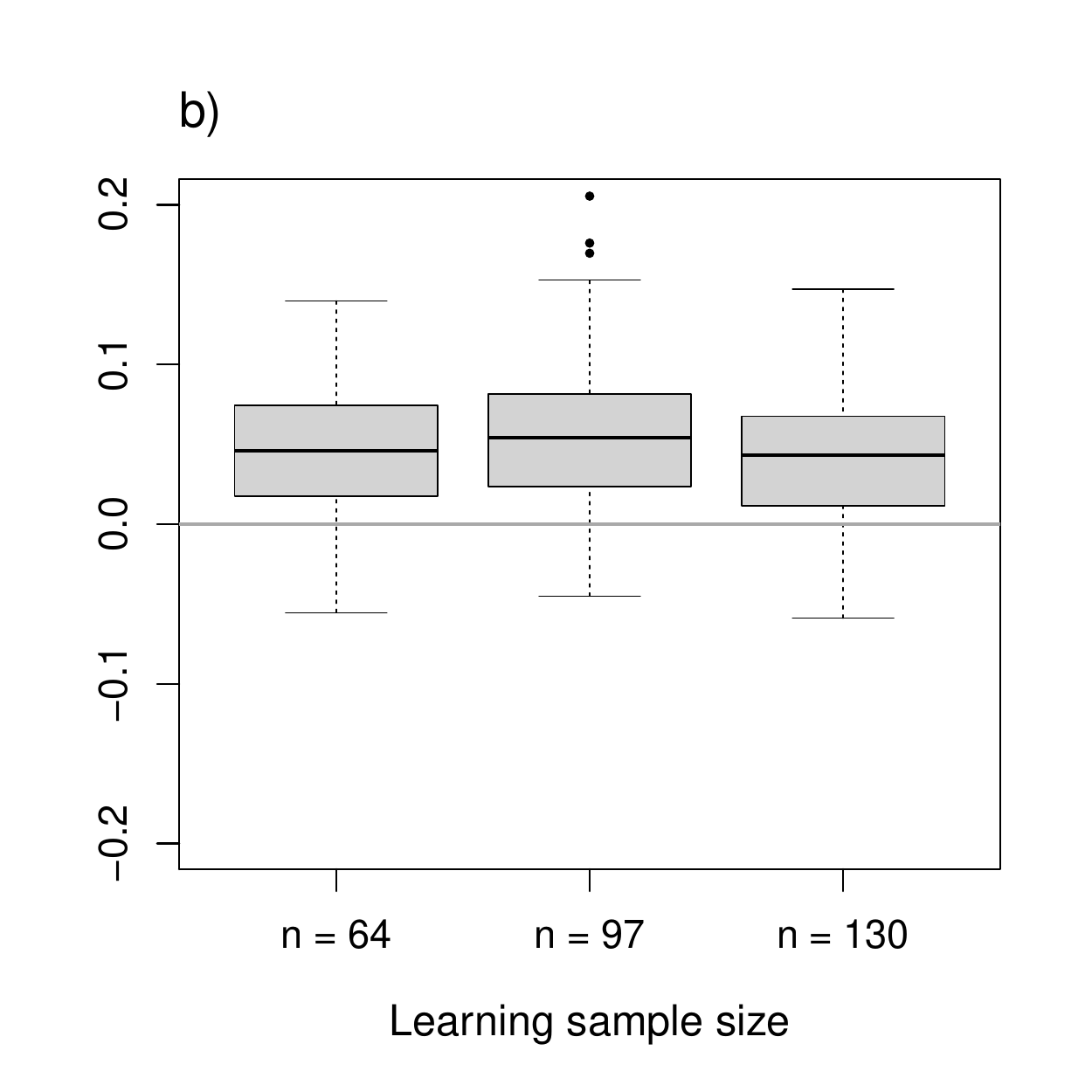}
	\caption{Boxplots of the performance differences between $C$-based and log-rank-based splitting in RSF for the breast cancer data (\citealp{Desmedt2007}). For all learning sample sizes, predictive performance was measured by calculating Harrell's $C$ from test data sets of size $n=66$ each. a) Data analysis using the five clinical variables only, b) low-dimensional data analysis using clinical and gene expression data.} \label{Fig:Des1}
\end{figure}

\begin{figure}[t!]
	\centering \vspace{-.5cm}
	\includegraphics[scale = 0.5]{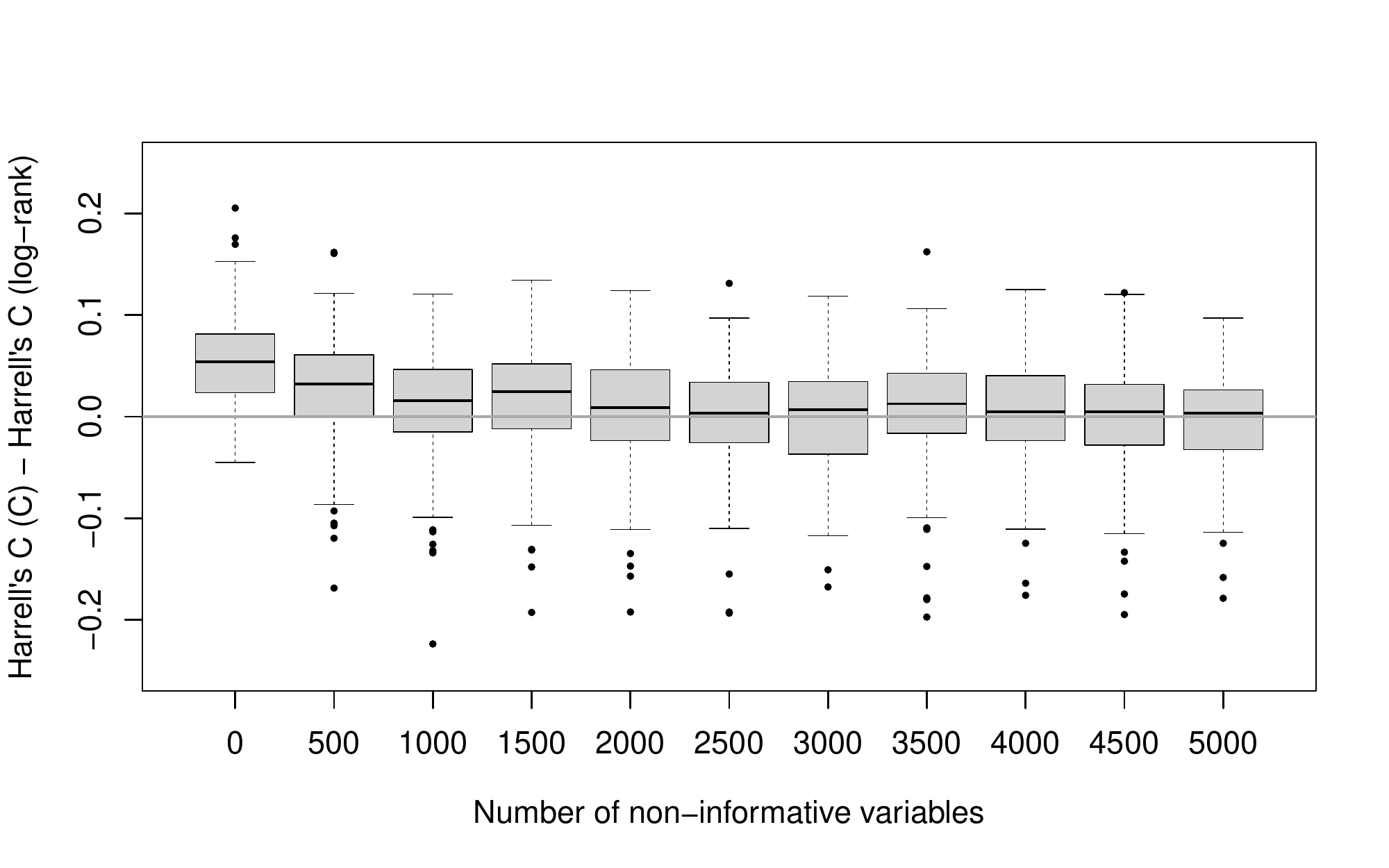}
	\caption{Boxplots of the performance differences between $C$-based and log-rank-based splitting in RSF for the high-dimensional analysis of the breast cancer data (\citealp{Desmedt2007}). Predictive performance was measured by calculating Harrell's $C$ from test data sets of size $n=66$ each. Non-informative variables ($p_{\mathrm{add}} = 1000, 1500, 2000, \ldots , 5000$) were added to the clinical variables and the gene expression variables.} \label{Fig:Des2}
\end{figure}

Only a small increase in RSF prediction accuracy was observed when log-rank splitting was replaced by $C$-based splitting and when the five clinical variables were used (see Figure \ref{Fig:Des1}.a), where all boxplots are close to zero.). This finding agrees well with the results of simulation study $2$.b), as three of the five clinical variables were categorical (cf.\@ Figure \ref{fig:simuEvalDiscrete}). In contrast, when the continuous gene expressions were added to the five clinical variables (Figure \ref{Fig:Des1}.b)), prediction accuracy was substantially higher for RSF using $C$-based splitting compared to RSF using log-rank splitting. As seen from Figure \ref{Fig:Des1}.b), even the lower quartiles of the differences between $C$-based and log-rank splitting were distinctly larger than zero. This result was observed for all learning sample sizes. In absolute value, median $C$ estimates obtained from $C$-based splitting were 0.61/0.61/0.60 in the five-variable clinical model and 0.67/0.71/0.73 in the combined model using both clinical and gene expression data (learning sample sizes $n = 64 / 97 /130$, respectively). The largest improvement obtained from $C$-based splitting was seen in the combined model with $n = 130$. In this case, the median $C$ estimate obtained from log-rank splitting was $0.685$, and the median improvement obtained from $C$-based splitting was $0.054$ ($95\%$ confidence interval $[0.039, 0.063]$), corresponding to a median performance gain of approximately~$8\%$. With increasing numbers of noise variables (Figure \ref{Fig:Des2}), the performance of $C$-based splitting decreased relative to log-rank-based splitting.

\begin{figure}[t!]
	\centering \vspace{-0.8cm}
	\includegraphics[scale = 0.55]{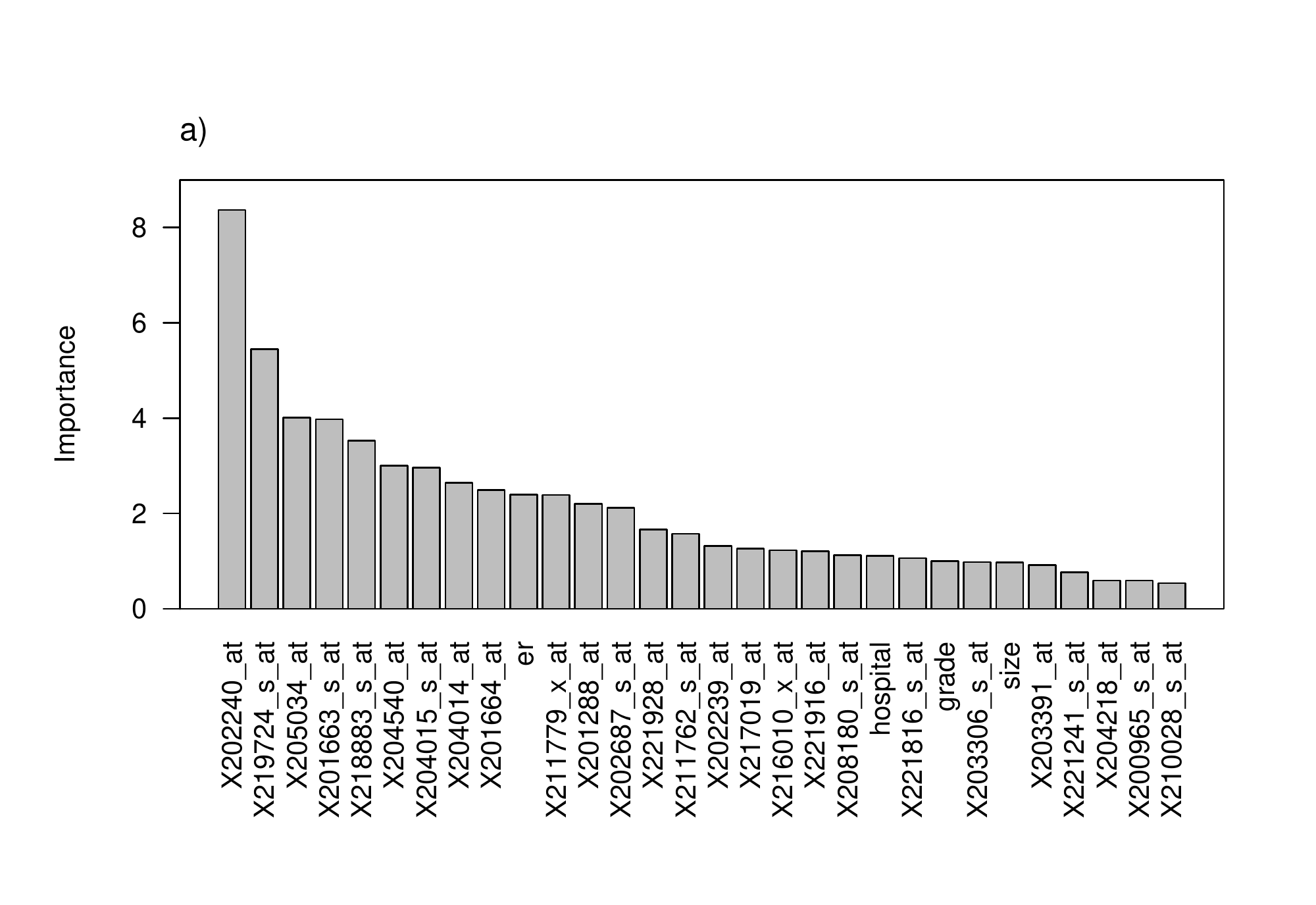} \\[-1.1cm]
	\includegraphics[scale = 0.55]{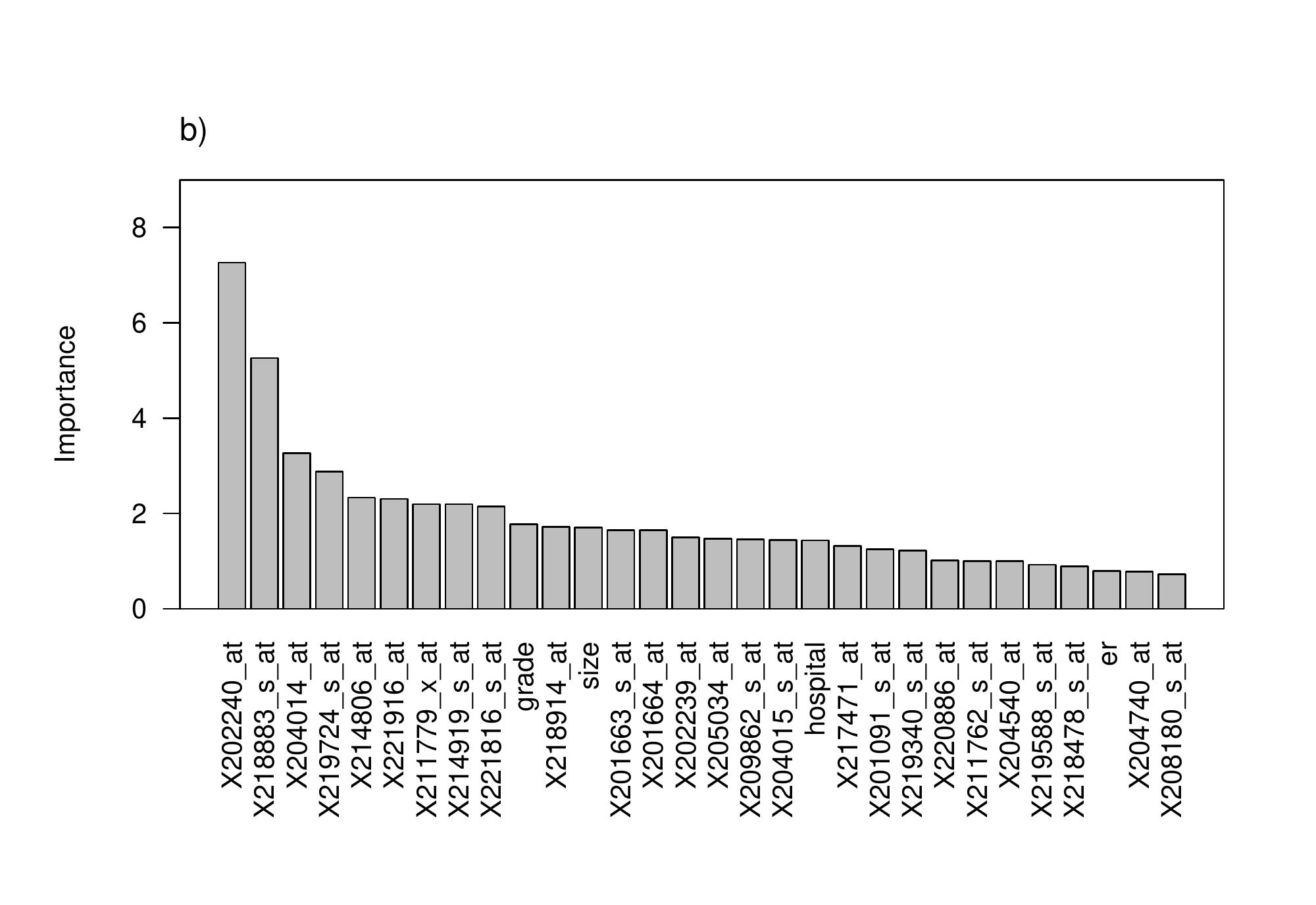} \vspace{-0.4cm}
	\caption{Scaled permutation-based importance scores for the $30$ most important clinical or gene expression variables in the RSF analysis of the breast cancer data (\citealp{Desmedt2007}). a) $C$-based splitting, b) log-rank splitting.} \label{fig:scores}
\end{figure}

In addition to the evaluation of RSF prediction accuracy, we also calculated permutation-based variable importance scores from the complete data set with $n=196$ observations. Figure \ref{fig:scores} shows that $C$-based and log-rank-based splitting yielded substantial differences in the permutation importance and the order of the importance of the predictor variables. For example, although both splitting rules assigned the highest importance score to gene 202240$\_$at, only 19 of the 30 most important `log-rank' predictors were also contained in the respective set of `$C$-based' predictors. Importantly, estrogen receptor status was the $10^{th}$ most important variable in the set of $C$-based predictors, while it was only ranked \#$28$ in the list of the most important predictor variables when log-rank splitting was used. Hospital location was among the most important predictor variables in both approaches, suggesting that the survival times showed considerable geographic variation.

Note that the main purpose of Figure \ref{fig:scores} is to demonstrate that the results of $C$-based splitting have a meaningful interpretation in regard to the inclusion of clinically relevant predictor variables. Additional simulation studies would be necessary to systematically investigate the relationship between permutation-based variable importance and $C$-based splitting. Also, there are various alternatives to permutation-based variable importance (e.g., \citealp{Ishwaran2011}), which might be used for additional variable selection purposes in high-dimensional settings. Investigating the behavior of these measures in RSF with $C$-based splitting may constitute an issue of future research.

\subsection{Data analysis 2 -- survival in patients with diffuse large B-cell lymphoma}
\label{s42r}
\begin{figure}[t!]
	\centering \vspace{-0.7cm}
	\includegraphics[scale = 0.5]{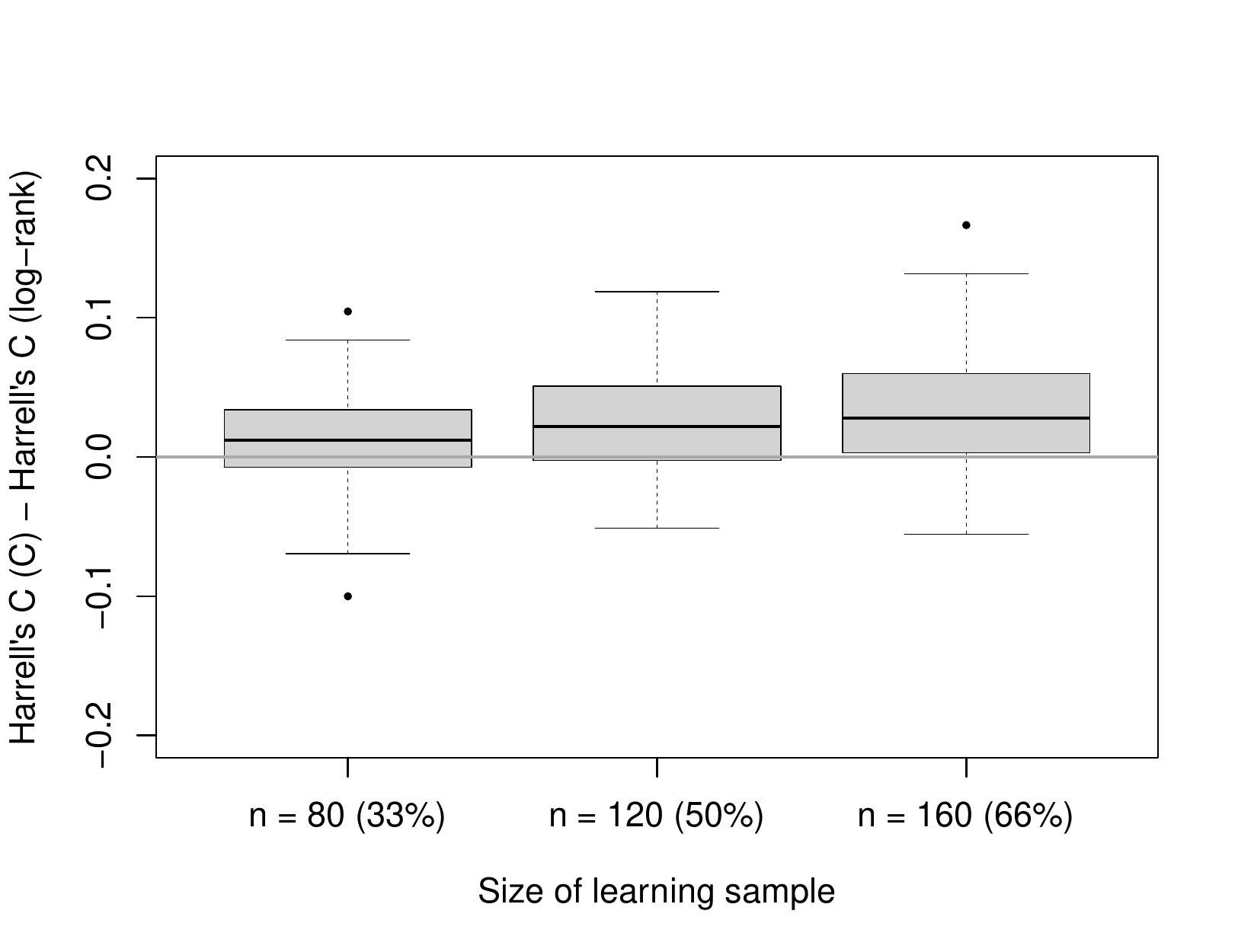}\vspace{-0.2cm}
	\caption{Boxplots of the performance differences between $C$-based and log-rank-based splitting in RSF for the high-dimensional diffuse large B-cell lymphoma data (\citealp{Rosenwald2002}). Predictive performance was measured by calculating Harrell's $C$ from evaluation data sets of size $n=80$ each.} \label{fig:rosenwald}
\end{figure}

Figure \ref{fig:rosenwald} shows the main results of the RSF analysis for the diffuse large B-cell lymphoma gene expression data. A substantial gain was observed in RSF performance when log-rank splitting was replaced by $C$-based splitting, as all boxplots in Figure \ref{fig:rosenwald} show a notable upward shift. This performance gain was observed for all sizes of the learning data.
In absolute value, median $C$ estimates obtained from $C$-based splitting were 0.60/0.61/0.62 (learning sample sizes $n = 80/120/160$, respectively). The largest improvement obtained from $C$-based splitting was seen in the model with $n = 160$. In this case, the median $C$ estimate obtained from log-rank splitting was $0.584$, and the median improvement obtained from $C$-based splitting was $0.028$ ($95\%$ confidence interval $[0.018, 0.041]$), corresponding to a median performance gain of approximately~$5\%$.

\section{Discussion}	\label{se:Discussion}

Random forests are an established machine learning technique in medical research \citep{Foster2011, Bacauskiene2012, Englund2012, Casanova2014, Calderoni2015}. In survival analysis, RSF are not only a powerful tool for risk prediction but are also a nonparametric alternative to Cox regression in situations where the proportional hazards assumption is violated \citep{Omurlu2009} or when the functional form of the predictor effects on survival is misspecified.

In this work, we have proposed Harrell's $C$ as an alternative split criterion to log-rank splitting for RSF. Conceptually, this approach corresponds to a unified analysis strategy, in which the split criterion is identical to the evaluation criterion of interest. Similar strategies have previously been developed for other statistical learning methods. For example, \cite{Mayr2014} proposed a gradient boosting algorithm for direct optimization of the concordance probability, implying that the same performance criterion was used for both model building and evaluation. In a similar way, several authors proposed support vector machines for censored data \citep{VanBelle2011,Polsterl2015}.

Based on our analytical and numerical results, we identified three situations in which RSF performance improved when Harrell's $C$ was used for node splitting instead of the log-rank statistic. First, $C$-based splitting performed better than log-rank splitting in scenarios where the signal-to-noise ratio was high in the data. This finding can be explained by the more unbalanced children nodes obtained from log-rank splitting. In the CART literature, a tendency towards unbalanced splits is referred to as `end-cut preference' (ECP) and has been a subject of debate for long \citep{Breiman1984}. In recent work, \cite{Ishwaran2015} provided an in-depth analysis of the ECP phenomenon and discussed its consequences for the performance of random forests. In particular, Ishwaran showed that split rules with ECP are desirable for trees with a small minimum node size when a non-informative predictor variable is considered for splitting. This is because end-cut splits conserve the sample size and therefore allow trees to `recover' from bad splits resulting from the splitting of a non-informative predictor variable. In the light of these findings, it is expected that split criteria with ECP increase the performance of RSF when applied to noisy data. Log-rank splitting, which tends to produce more unbalanced splits than Harrell's~$C$, is therefore expected to result in an improved RSF performance when the percentage of non-informative predictors is high. This explains the good performance of the log-rank statistic under the noisy scenarios compared to Harrell's~$C$.

Second, $C$-based splitting resulted in substantial performance gains when the number of informative {\em continuous} predictor variables was large compared to the number of {\em categorical} predictor variables. Conversely, $C$-based and log-rank-based splitting resulted in little difference in RSF performance when all predictor variables were categorical. This result is explained by the fact that differences in threshold selection become most apparent in the presence of continuous predictor variables.

Third, we observed that $C$-based splitting tended to outperform log-rank splitting in data sets with a high censoring rate. This result confirmed earlier findings by \cite{Ishwaran2008} who stated that `the performance of [log-rank-based] RF regression depended strongly on the censoring rate', with the prediction accuracy of RSF being `poor' in high-censoring scenarios.

In summary, our results show that Harrell's $C$ as split criterion improves the performance of RSF in a wide range of settings. We recommend the use of $C$-based splitting in smaller scale clinical studies and the use of log-rank splitting in large-scale `omics' studies.

\section*{Acknowledgements}

The work of MS has been supported by Deutsche Forschungsgemeinschaft (DFG), grant SCHM 2966/1-2.
The work of AZ has been supported by the European Union Seventh Framework Programme (FP7/2007--2013) under grant
agreement  No.\  HEALTH-F2-2011-278913 (BiomarCaRE) and by the German Center for Cardiovascular Research (DZHK), Standort Hamburg/Kiel/L\"ubeck, L\"ubeck, Germany.

\section*{References}

\bibliography{literature/litSM}

\end{document}